\documentclass[10pt,twocolumn,letterpaper]{article}

\usepackage{cvpr}
\usepackage{times}
\usepackage{epsfig}
\usepackage{graphicx}
\usepackage{amsmath}
\usepackage{amssymb}
\usepackage{array}
\usepackage{enumitem}
\usepackage{stfloats}

\usepackage[pagebackref=true,breaklinks=true,letterpaper=true,colorlinks,bookmarks=false]{hyperref}

\cvprfinalcopy 

\def\cvprPaperID{1429} 

\ifcvprfinal\fi

\newcommand{\name}{fully convolutional network}
\newcommand{\Name}{Fully convolutional network}
\newcommand{\NAME}{FCN}
\newcommand{\minisection}[1]{\textbf{#1}\hspace{0.3em}}

\newcommand{\metr}[1]{\parbox[b]{0.2in}{#1}}
\newcommand{\pixacc}{\metr{pixel\\acc.}}
\newcommand{\classacc}{\metr{mean\\acc.}}
\newcommand{\meanIU}{\metr{mean\\IU}}
\newcommand{\fwIU}{\metr{f.w.\\IU}}

\begin{document}

\title{Fully Convolutional Networks for Semantic Segmentation}

\author{Jonathan Long\thanks{Authors contributed equally}\hspace{2em}
    Evan Shelhamer\footnotemark[1]\hspace{2em}
    Trevor Darrell\\
    UC Berkeley\\
    {\tt\small \{jonlong,shelhamer,trevor\}@cs.berkeley.edu}
}

\maketitle

\begin{abstract}
Convolutional networks are powerful visual models that yield hierarchies of features.
We show that convolutional networks by themselves, trained end-to-end, pixels-to-pixels, exceed the state-of-the-art in semantic segmentation.
Our key insight is to build ``fully convolutional'' networks that take input of arbitrary size and produce correspondingly-sized output with efficient inference and learning.
We define and detail the space of \name s, explain their application to spatially dense prediction tasks, and draw connections to prior models.
We adapt contemporary classification networks (AlexNet \cite{AlexNet}, the VGG net \cite{VGGNet}, and GoogLeNet \cite{GoogLeNet}) into \name s and transfer their learned representations by fine-tuning \cite{decaf} to the segmentation task.
We then define a novel architecture that combines semantic information from a deep, coarse layer with appearance information from a shallow, fine layer to produce accurate and detailed segmentations.
Our \name\ achieves state-of-the-art segmentation of PASCAL VOC (20\% relative improvement to 62.2\% mean IU on 2012), NYUDv2, and SIFT Flow, while inference takes less than one fifth of a second for a typical image.
\end{abstract}

\section{Introduction}


Convolutional networks are driving advances in recognition.
Convnets are not only improving for whole-image classification \cite{AlexNet, VGGNet, GoogLeNet}, but also making progress on local tasks with structured output.
These include advances in bounding box object detection \cite{OverFeat, RCNN, SPP}, part and keypoint prediction \cite{Ning, Jon}, and local correspondence \cite{Jon, BroxSIFT}.

The natural next step in the progression from coarse to fine inference is to make a prediction at every pixel.
Prior approaches have used convnets for semantic segmentation \cite{ning2005automatic, Ciresan, Farabet, Pinheiro, Bharath, Saurabh, N4}, in which each pixel is labeled with the class of its enclosing object or region, but with shortcomings that this work addresses.

We show that a \name\ (\NAME), trained end-to-end, pixels-to-pixels on semantic segmentation exceeds the state-of-the-art without further machinery.
To our knowledge, this is the first work to train \NAME s end-to-end (1) for pixelwise prediction and (2) from supervised pre-training.
Fully convolutional versions of existing networks predict dense outputs from arbitrary-sized inputs.
Both learning and inference are performed whole-image-at-a-time by dense feedforward computation and backpropagation.
In-network upsampling layers enable pixelwise prediction and learning in nets with subsampled pooling.

This method is efficient, both asymptotically and absolutely, and precludes the need for the complications in other works.
Patchwise training is common \cite{ning2005automatic, Ciresan, Farabet, Pinheiro, N4}, but lacks the efficiency of fully convolutional training.
Our approach does not make use of pre- and post-processing complications, including superpixels \cite{Farabet, Bharath}, proposals \cite{Bharath, Saurabh}, or post-hoc refinement by random fields or local classifiers \cite{Farabet, Bharath}.
Our model transfers recent success in classification \cite{AlexNet, VGGNet, GoogLeNet} to dense prediction by reinterpreting classification nets as fully convolutional and fine-tuning from their learned representations.
In contrast, previous works have applied small convnets without supervised pre-training \cite{Farabet, Pinheiro, ning2005automatic}.

\begin{figure}
\centering
\includegraphics[width=0.45\textwidth]{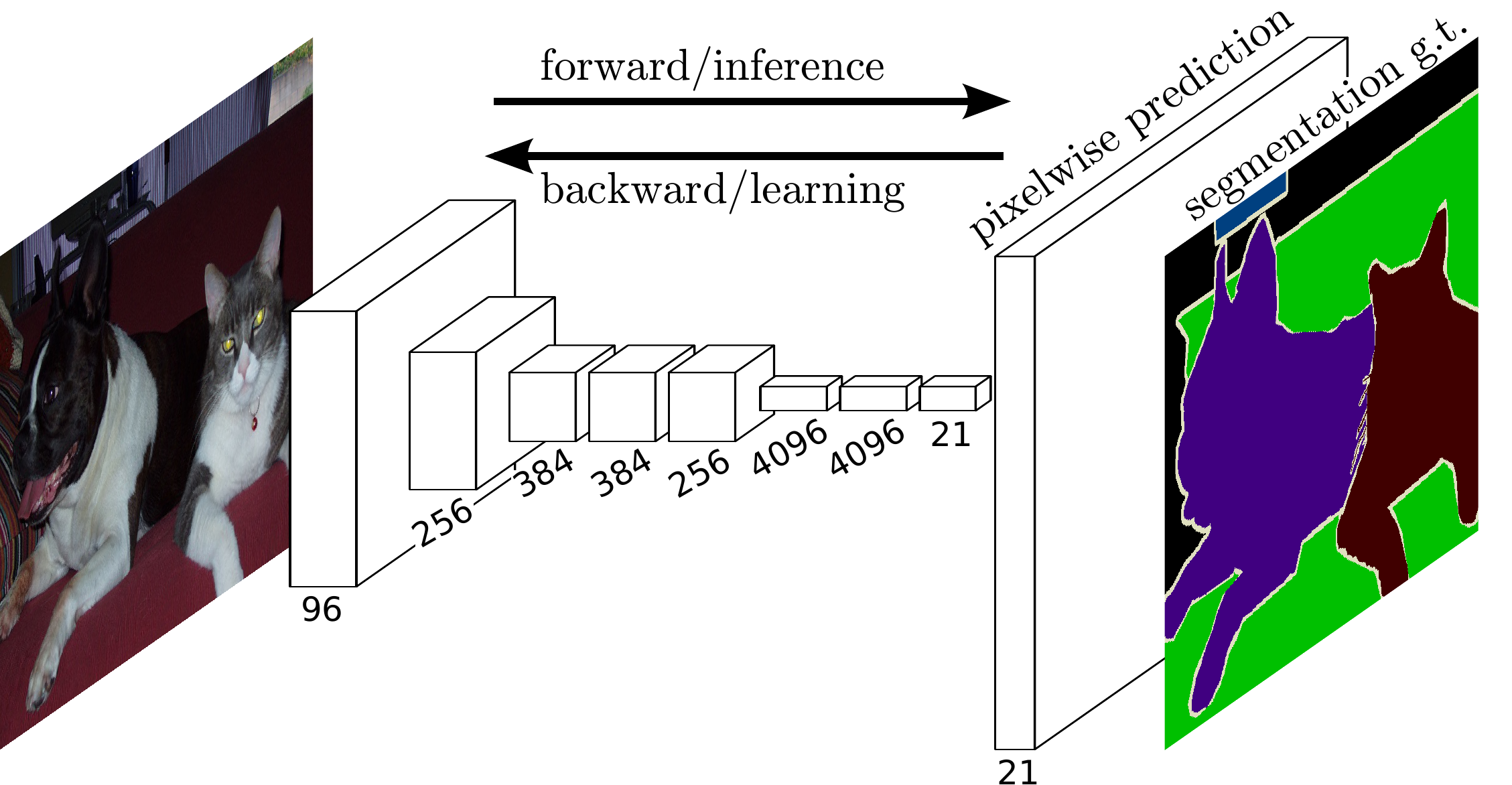}
\caption{
Fully convolutional networks can efficiently learn to make dense predictions for per-pixel tasks like semantic segmentation.
}
\label{fig:model}
\end{figure}

Semantic segmentation faces an inherent tension between semantics and location: global information resolves what while local information resolves where.
Deep feature hierarchies jointly encode location and semantics in a local-to-global pyramid.
We define a novel ``skip'' architecture to combine deep, coarse, semantic information and shallow, fine, appearance information in Section \ref{sec:skip} (see Figure \ref{fig:nets}).

In the next section, we review related work on deep classification nets, \NAME s, and recent approaches to semantic segmentation using convnets.
The following sections explain \NAME\ design and dense prediction tradeoffs, introduce our architecture with in-network upsampling and multi-layer combinations, and describe our experimental framework.
Finally, we demonstrate state-of-the-art results on PASCAL VOC 2011-2, NYUDv2, and SIFT Flow.

\section{Related work}

Our approach draws on recent successes of deep nets for image classification \cite{AlexNet, VGGNet, GoogLeNet} and transfer learning \cite{decaf,ZF}.
Transfer was first demonstrated on various visual recognition tasks \cite{decaf,ZF}, then on detection, and on both instance and semantic segmentation in hybrid proposal-classifier models \cite{RCNN, Bharath, Saurabh}.
We now re-architect and fine-tune classification nets to direct, dense prediction of semantic segmentation.
We chart the space of \NAME s and situate prior models, both historical and recent, in this framework.


\minisection{\Name s}
To our knowledge, the idea of extending a convnet to arbitrary-sized inputs first appeared in Matan \etal \cite{Matan}, which extended the classic LeNet \cite{LeNet} to recognize strings of digits.
Because their net was limited to one-dimensional input strings, Matan \etal used Viterbi decoding to obtain their outputs.
Wolf and Platt \cite{wolf1994postal} expand convnet outputs to 2-dimensional maps of detection scores for the four corners of postal address blocks.
Both of these historical works do inference and learning fully convolutionally for detection.
Ning \etal \cite{ning2005automatic} define a convnet for coarse multiclass segmentation of \emph{C. elegans} tissues with fully convolutional inference. 

Fully convolutional computation has also been exploited in the present era of many-layered nets.
Sliding window detection by Sermanet \etal \cite{OverFeat}, semantic segmentation by Pinheiro and Collobert \cite{Pinheiro}, and image restoration by Eigen \etal \cite{Eigenrestore} do fully convolutional inference. 
Fully convolutional training is rare, but used effectively by Tompson \etal \cite{Tompson} to learn an end-to-end part detector and spatial model for pose estimation, although they do not exposit on or analyze this method.

Alternatively, He \etal \cite{SPP} discard the non-convolutional portion of classification nets to make a feature extractor.
They combine proposals and spatial pyramid pooling to yield a localized, fixed-length feature for classification.
While fast and effective, this hybrid model cannot be learned end-to-end.

\minisection{Dense prediction with convnets}
Several recent works have applied convnets to dense prediction problems,
including semantic segmentation by Ning \etal \cite{ning2005automatic}, Farabet \etal \cite{Farabet}, and Pinheiro and Collobert \cite{Pinheiro};
boundary prediction for electron microscopy by Ciresan \etal \cite{Ciresan} and for natural images by a hybrid neural net/nearest neighbor model by Ganin and Lempitsky \cite{N4};
and image restoration and depth estimation by Eigen \etal \cite{Eigenrestore, Eigendepth}.
Common elements of these approaches include
\begin{itemize}[noitemsep,topsep=0pt,parsep=0pt,partopsep=0pt]
  \item small models restricting capacity and receptive fields;
  \item patchwise training \cite{ning2005automatic, Ciresan, Farabet, Pinheiro, N4};
  \item post-processing by superpixel projection, random field regularization, filtering, or local classification \cite{Farabet, Ciresan, N4};
  \item input shifting and output interlacing for dense output \cite{Pinheiro, N4} as introduced by OverFeat \cite{OverFeat};
  \item multi-scale pyramid processing \cite{Farabet, Pinheiro, N4};
  \item saturating $\tanh$ nonlinearities \cite{Farabet, Eigenrestore, Pinheiro}; and
  \item ensembles \cite{Ciresan, N4},
\end{itemize}
whereas our method does without this machinery. However, we do study patchwise training \ref{sec:patches} and ``shift-and-stitch'' dense output \ref{sec:shifting} from the perspective of \NAME s.
We also discuss in-network upsampling \ref{sec:upsampling}, of which the fully connected prediction by Eigen \etal \cite{Eigendepth} is a special case.

Unlike these existing methods, we adapt and extend deep classification architectures, using image classification as supervised pre-training, and fine-tune fully convolutionally to learn simply and efficiently from whole image inputs and whole image ground thruths.

Hariharan \etal \cite{Bharath} and Gupta \etal \cite{Saurabh} likewise adapt deep classification nets to semantic segmentation, but do so in hybrid proposal-classifier models.
These approaches fine-tune an R-CNN system \cite{RCNN} by sampling bounding boxes and/or region proposals for detection, semantic segmentation, and instance segmentation.
Neither method is learned end-to-end.

They achieve state-of-the-art results on PASCAL VOC segmentation and NYUDv2 segmentation respectively, so we directly compare our standalone, end-to-end \NAME\ to their semantic segmentation results in Section \ref{sec:results}.

\section{\Name s}
\label{sec:fc}

Each layer of data in a convnet is a three-dimensional array of size $h \times w \times d$, where $h$ and $w$ are spatial dimensions, and $d$ is the feature or channel dimension.
The first layer is the image, with pixel size $h \times w$, and $d$ color channels.
Locations in higher layers correspond to the locations in the image they are path-connected to, which are called their \emph{receptive fields}.

Convnets are built on translation invariance.
Their basic components (convolution, pooling, and activation functions) operate on local input regions, and depend only on \emph{relative} spatial coordinates.
Writing $\mathbf x_{ij}$ for the data vector at location $(i, j)$ in a particular layer, and $\mathbf y_{ij}$ for the following layer, these functions compute outputs $\mathbf y_{ij}$ by
\[
\mathbf y_{ij} = f_{ks}\left(
\{\mathbf x_{si + \delta i, sj + \delta j}\}_{0 \leq \delta i,
\delta j \leq k}\right)
\]
where $k$ is called the kernel size, $s$ is the stride or subsampling factor, and $f_{ks}$ determines the layer type: a matrix multiplication for convolution or average pooling, a spatial max for max pooling, or an elementwise nonlinearity for an activation function, and so on for other types of layers.

This functional form is maintained under composition, with kernel size and stride obeying the transformation rule
\[
f_{ks} \circ g_{k's'} = (f \circ g)_{k'+(k-1)s', ss'}.
\]
While a general deep net computes a general nonlinear function, a net with only layers of this form computes a nonlinear \emph{filter}, which we call a \emph{deep filter} or \emph{fully convolutional network}.
An \NAME\ naturally operates on an input of any size, and produces an output of corresponding (possibly resampled) spatial dimensions.

A real-valued loss function composed with an \NAME\ defines a task.
If the loss function is a sum over the spatial dimensions of the final layer, $\ell(\mathbf x; \theta) = \sum_{ij} \ell'(\mathbf x_{ij}; \theta)$, its gradient will be a sum over the gradients of each of its spatial components.
Thus stochastic gradient descent on $\ell$ computed on whole images will be the same as stochastic gradient descent on $\ell'$, taking all of the final layer receptive fields as a minibatch.

When these receptive fields overlap significantly, both feedforward computation \emph{and} backpropagation are much more efficient when computed layer-by-layer over an entire image instead of independently patch-by-patch.

We next explain how to convert classification nets into fully convolutional nets that produce coarse output maps.
For pixelwise prediction, we need to connect these coarse outputs back to the pixels.
Section \ref{sec:shifting} describes a trick that OverFeat \cite{OverFeat} introduced for this purpose.
We gain insight into this trick by reinterpreting it as an equivalent network modification.
As an efficient, effective alternative, we introduce deconvolution layers for upsampling in Section \ref{sec:upsampling}.
In Section \ref{sec:patches} we consider training by patchwise sampling, and give evidence in Section \ref{sec:training} that our whole image training is faster and equally effective.

\subsection{Adapting classifiers for dense prediction}
\label{sec:adapting}

\begin{figure}
\includegraphics[width=0.45\textwidth]{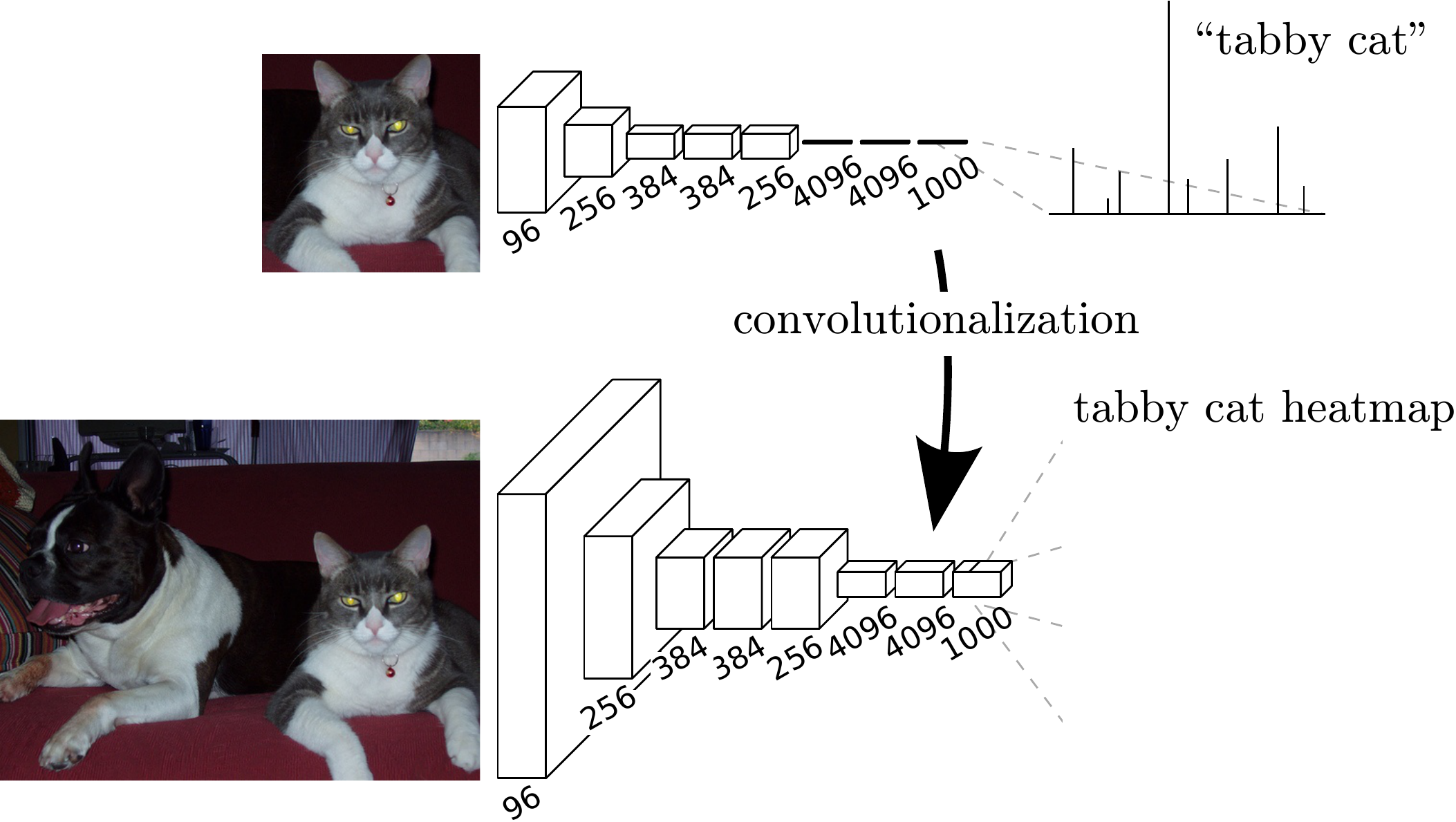}
\caption{
Transforming fully connected layers into convolution layers enables a classification net to output a heatmap.
Adding layers and a spatial loss (as in Figure \ref{fig:model}) produces an efficient machine for end-to-end dense learning.
}
\label{fig:alex-evo}
\end{figure}

Typical recognition nets, including LeNet \cite{LeNet}, AlexNet \cite{AlexNet}, and its deeper successors \cite{VGGNet, GoogLeNet}, ostensibly take fixed-sized inputs and produce nonspatial outputs.
The fully connected layers of these nets have fixed dimensions and throw away spatial coordinates.
However, these fully connected layers can also be viewed as convolutions with kernels that cover their entire input regions.
Doing so casts them into \name s that take input of any size and output classification maps.
This transformation is illustrated in Figure \ref{fig:alex-evo}.
(By contrast, nonconvolutional nets, such as the one by Le \etal \cite{Le}, lack this capability.)

Furthermore, while the resulting maps are equivalent to the evaluation of the original net on particular input patches, the computation is highly amortized over the overlapping regions of those patches.
For example, while AlexNet takes $1.2$ ms (on a typical GPU) to produce the classification scores of a $227 \times 227$ image, the fully convolutional version takes $22$ ms to produce a $10 \times 10$ grid of outputs from a $500 \times 500$ image, which is more than $5$ times faster than the na\"ive approach\footnote{Assuming efficient batching of single image inputs.
The classification scores for a single image by itself take 5.4 ms to produce, which is nearly $25$ times slower than the fully convolutional version.}.

The spatial output maps of these convolutionalized models make them a natural choice for dense problems like semantic segmentation.
With ground truth available at every output cell, both the forward and backward passes are straightforward, and both take advantage of the inherent computational efficiency (and aggressive optimization) of convolution.

The corresponding backward times for the AlexNet example are $2.4$ ms for a single image and $37$ ms for a fully convolutional $10 \times 10$ output map, resulting in a speedup similar to that of the forward pass.
This dense backpropagation is illustrated in Figure \ref{fig:model}.

While our reinterpretation of classification nets as fully convolutional yields output maps for inputs of any size, the output dimensions are typically reduced by subsampling.
The classification nets subsample to keep filters small and computational requirements reasonable.
This coarsens the output of a fully convolutional version of these nets, reducing it from the size of the input by a factor equal to the pixel stride of the receptive fields of the output units.



\subsection{Shift-and-stitch is filter rarefaction}
\label{sec:shifting}

Input shifting and output interlacing is a trick that yields dense predictions from coarse outputs without interpolation, introduced by OverFeat \cite{OverFeat}.
If the outputs are downsampled by a factor of $f$, the input is shifted (by left and top padding) $x$ pixels to the right and $y$ pixels down, once for every value of $(x, y) \in \{0,\ldots,f-1\} \times \{0, \ldots, f-1\}$.
These $f^2$ inputs are each run through the convnet, and the outputs are interlaced so that the predictions correspond to the pixels at the \emph{centers} of their receptive fields.

Changing only the filters and layer strides of a convnet can produce the same output as this shift-and-stitch trick.
Consider a layer (convolution or pooling) with input stride $s$, and a following convolution layer with filter weights $f_{ij}$ (eliding the feature dimensions, irrelevant here).
Setting the lower layer's input stride to $1$ upsamples its output by a factor of $s$, just like shift-and-stitch.
However, convolving the original filter with the upsampled output does not produce the same result as the trick, because the original filter only sees a reduced portion of its (now upsampled) input.
To reproduce the trick, rarefy the filter by enlarging it as
\[
f'_{ij} = \left\{
\begin{tabular}{ll}
$f_{i/s, j/s}$ & if $s$ divides both $i$ and $j$; \\
$0$ & otherwise,
\end{tabular}
\right.
\]
(with $i$ and $j$ zero-based).
Reproducing the full net output of the trick involves repeating this filter enlargement layer-by-layer until all subsampling is removed.

Simply decreasing subsampling within a net is a tradeoff: the filters see finer information, but have smaller receptive fields and take longer to compute.
We have seen that the shift-and-stitch trick is another kind of tradeoff: the output is made denser without decreasing the receptive field sizes of the filters, but the filters are prohibited from accessing information at a finer scale than their original design.

Although we have done preliminary experiments with shift-and-stitch, we do not use it in our model.
We find learning through upsampling, as described in the next section, to be more effective and efficient, especially when combined with the skip layer fusion described later on.

\subsection{Upsampling is backwards strided convolution}
\label{sec:upsampling}

Another way to connect coarse outputs to dense pixels is interpolation.
For instance, simple bilinear interpolation computes each output $y_{ij}$ from the nearest four inputs by a linear map that depends only on the relative positions of the input and output cells.

In a sense, upsampling with factor $f$ is convolution with a \emph{fractional} input stride of $1/f$.
So long as $f$ is integral, a natural way to upsample is therefore \emph{backwards convolution} (sometimes called \emph{deconvolution}) with an \emph{output} stride of $f$.
Such an operation is trivial to implement, since it simply reverses the forward and backward passes of convolution.
Thus upsampling is performed in-network for end-to-end learning by backpropagation from the pixelwise loss.

Note that the deconvolution filter in such a layer need not be fixed (e.g., to bilinear upsampling), but can be learned.
A stack of deconvolution layers and activation functions can even learn a nonlinear upsampling.

In our experiments, we find that in-network upsampling is fast and effective for learning dense prediction.
Our best segmentation architecture uses these layers to learn to upsample for refined prediction in Section \ref{sec:skip}.

\subsection{Patchwise training is loss sampling}
\label{sec:patches}

In stochastic optimization, gradient computation is driven by the training distribution.
Both patchwise training and fully-convolutional training can be made to produce any distribution, although their relative computational efficiency depends on overlap and minibatch size.
Whole image fully convolutional training is identical to patchwise training where each batch consists of all the receptive fields of the units below the loss for an image (or collection of images).
While this is more efficient than uniform sampling of patches, it reduces the number of possible batches.
However, random selection of patches within an image may be recovered simply.
Restricting the loss to a randomly sampled subset of its spatial terms (or, equivalently applying a DropConnect mask \cite{dropconnect} between the output and the loss) excludes patches from the gradient computation.

If the kept patches still have significant overlap, fully convolutional computation will still speed up training.
If gradients are accumulated over multiple backward passes, batches can include patches from several images.%
\footnote{Note that not every possible patch is included this way, since the receptive fields of the final layer units lie on a fixed, strided grid.
However, by shifting the image left and down by a random value up to the stride, random selection from all possible patches may be recovered.}

Sampling in patchwise training can correct class imbalance \cite{ning2005automatic, Farabet, Ciresan} and mitigate the spatial correlation of dense patches \cite{Pinheiro, Bharath}.
In fully convolutional training, class balance can also be achieved by weighting the loss, and loss sampling can be used to address spatial correlation.

We explore training with sampling in Section \ref{sec:training}, and do not find that it yields faster or better convergence for dense prediction. Whole image training is effective and efficient.

\section{Segmentation Architecture}
\label{sec:arch}

We cast ILSVRC classifiers into \NAME s and augment them for dense prediction with in-network upsampling and a pixelwise loss.
We train for segmentation by fine-tuning.
Next, we build a novel skip architecture that combines coarse, semantic and local, appearance information to refine prediction.

For this investigation, we train and validate on the PASCAL VOC 2011 segmentation challenge \cite{PASCAL}.
We train with a per-pixel multinomial logistic loss and validate with the standard metric of mean pixel intersection over union, with the mean taken over all classes, including background.
The training ignores pixels that are masked out (as ambiguous or difficult) in the ground truth.

\subsection{From classifier to dense \NAME}
\label{sec:base}

\begin{figure*}[bp]
\includegraphics[width=\textwidth]{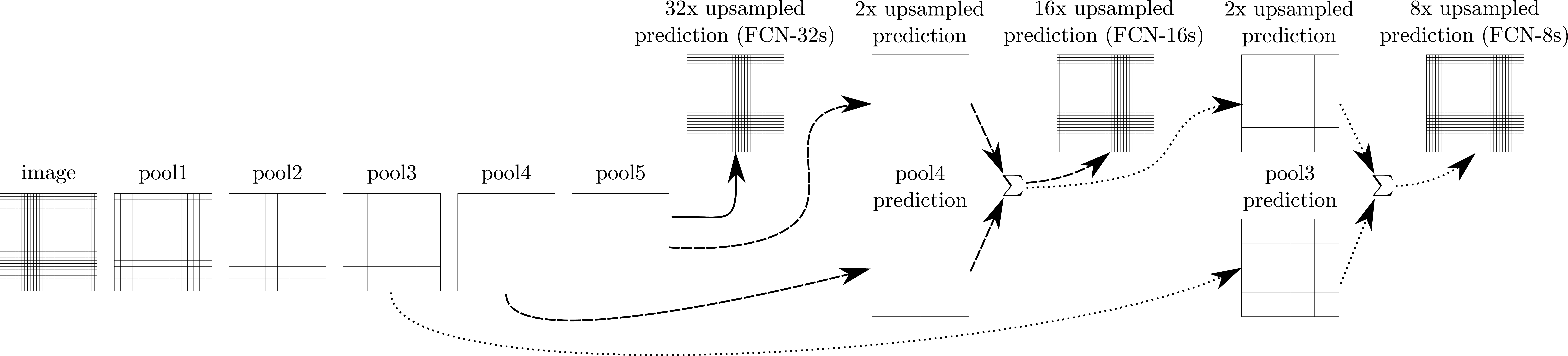}
\caption{
Our DAG nets learn to combine coarse, high layer information with fine, low layer information.
Layers are shown as grids that reveal relative spatial coarseness.
Only pooling and prediction layers are shown; intermediate convolution layers (including our converted fully connected layers) are omitted.
Solid line (FCN-32s): Our single-stream net, described in Section \ref{sec:base}, upsamples stride 32 predictions back to pixels in a single step.
Dashed line (FCN-16s): Combining predictions from both the final layer and the \texttt{pool4} layer, at stride 16, lets our net predict finer details, while retaining high-level semantic information.
Dotted line (FCN-8s): Additional predictions from \texttt{pool3}, at stride 8, provide further precision.
}
\label{fig:nets}
\end{figure*}

We begin by convolutionalizing proven classification architectures as in Section \ref{sec:fc}.
We consider the AlexNet\footnote{Using the publicly available \texttt{CaffeNet} reference model.} architecture \cite{AlexNet} that won ILSVRC12, as well as the VGG nets \cite{VGGNet} and the GoogLeNet%
\footnote{%
Since there is no publicly available version of GoogLeNet, we use our own reimplementation.
Our version is trained with less extensive data augmentation, and gets 68.5\% top-1 and 88.4\% top-5 ILSVRC accuracy.} \cite{GoogLeNet} which did exceptionally well in ILSVRC14.
We pick the VGG 16-layer net\footnote{Using the publicly available version from the Caffe model zoo.}, which we found to be equivalent to the 19-layer net on this task.
For GoogLeNet, we use only the final loss layer, and improve performance by discarding the final average pooling layer.
We decapitate each net by discarding the final classifier layer, and convert all fully connected layers to convolutions.
We append a $1 \times 1$ convolution with channel dimension $21$ to predict scores for each of the PASCAL classes (including background) at each of the coarse output locations, followed by a deconvolution layer to bilinearly upsample the coarse outputs to pixel-dense outputs as described in Section \ref{sec:upsampling}.
Table \ref{tab:base} compares the preliminary validation results along with the basic characteristics of each net.
We report the best results achieved after convergence at a fixed learning rate (at least 175 epochs).

Fine-tuning from classification to segmentation gave reasonable predictions for each net.
Even the worst model achieved $\sim 75\%$ of state-of-the-art performance.
The segmentation-equippped VGG net (FCN-VGG16) already appears to be state-of-the-art at 56.0 mean IU on val, compared to 52.6 on test \cite{Bharath}.
Training on extra data raises performance to 59.4 mean IU on a subset of val\footnotemark[7].
Training details are given in Section \ref{sec:training}.

Despite similar classification accuracy, our implementation of GoogLeNet did not match this segmentation result.

%
%

\begin{table}
\caption{
We adapt and extend three classification convnets to segmentation.
We compare performance by mean intersection over union on the validation set of PASCAL VOC 2011 and by inference time (averaged over 20 trials for a $500 \times 500$ input on an NVIDIA Tesla K40c).
We detail the architecture of the adapted nets as regards dense prediction: number of parameter layers, receptive field size of output units, and the coarsest stride within the net.
(These numbers give the best performance obtained at a fixed learning rate, not best performance possible.)
}
\setlength{\tabcolsep}{2pt}
\begin{tabular}{r|ccc}
& \parbox[b]{0.7in}{\centering FCN-AlexNet} & \parbox[b]{0.7in}{\centering FCN-VGG16} & \parbox[b]{0.9in}{\centering FCN-GoogLeNet\footnotemark[4]} \\
\hline
mean IU & 39.8 & \textbf{56.0} & 42.5 \\
forward time & 50 ms & 210 ms & 59 ms \\
conv.\ layers & 8 & 16 & 22 \\
parameters & 57M & 134M & 6M \\
rf size & 355 & 404 & 907 \\
max stride & 32 & 32 & 32
\end{tabular}
\label{tab:base}
\end{table}

\subsection{Combining what and where}
\label{sec:skip}

We define a new fully convolutional net (\NAME) for segmentation that combines layers of the feature hierarchy and refines the spatial precision of the output. See Figure \ref{fig:nets}.

\begin{figure}
\centering
\setlength{\tabcolsep}{1pt}
\begin{tabular}{cccc}
FCN-32s & FCN-16s & FCN-8s & Ground truth \\
\includegraphics[width=0.12\textwidth]{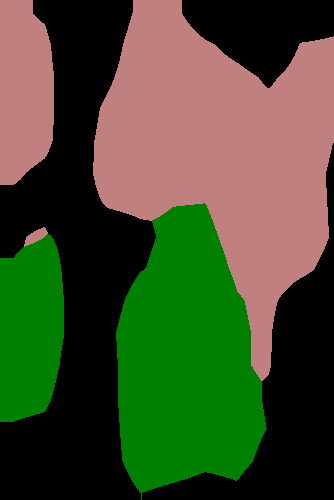} &
\includegraphics[width=0.12\textwidth]{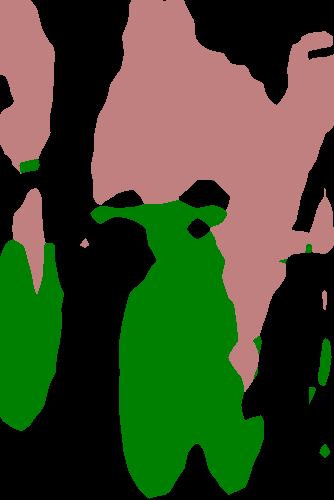} &
\includegraphics[width=0.12\textwidth]{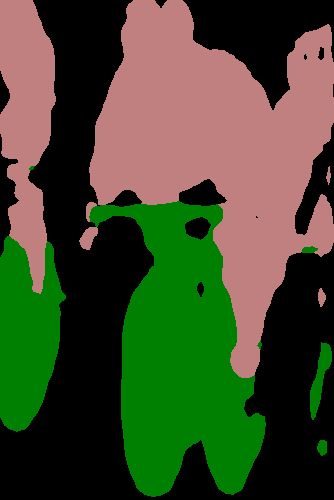} &
\includegraphics[width=0.12\textwidth]{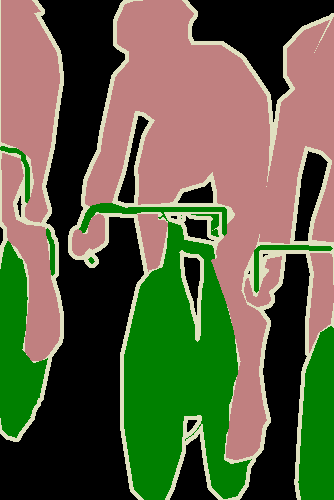}
\end{tabular}
\caption{
Refining fully convolutional nets by fusing information from layers with different strides improves segmentation detail.
The first three images show the output from our 32, 16, and 8 pixel stride nets (see Figure \ref{fig:nets}).
}
\label{fig:evo}
\end{figure}

While fully convolutionalized classifiers can be fine-tuned to segmentation as shown in \ref{sec:base}, and even score highly on the standard metric, their output is dissatisfyingly coarse (see Figure \ref{fig:evo}).
The 32 pixel stride at the final prediction layer limits the scale of detail in the upsampled output.

We address this by adding links that combine the final prediction layer with lower layers with finer strides.
This turns a line topology into a DAG, with edges that skip ahead from lower layers to higher ones (Figure \ref{fig:nets}).
As they see fewer pixels, the finer scale predictions should need fewer layers, so it makes sense to make them from shallower net outputs.
Combining fine layers and coarse layers lets the model make local predictions that respect global structure.
By analogy to the multiscale local jet of Florack \etal \cite{Florack}, we call our nonlinear local feature hierarchy the \emph{deep jet}.

We first divide the output stride in half by predicting from a 16 pixel stride layer.
We add a $1 \times 1$ convolution layer on top of \texttt{pool4} to produce additional class predictions.
We fuse this output with the predictions computed on top of \texttt{conv7} (convolutionalized \texttt{fc7}) at stride 32 by adding a $2\times$ upsampling layer
and summing\footnote{Max fusion made learning difficult due to gradient switching.} both predictions.
(See Figure \ref{fig:nets}).
We initialize the $2\times$ upsampling to bilinear interpolation, but allow the parameters to be learned as described in Section \ref{sec:upsampling}.
Finally, the stride 16 predictions are upsampled back to the image.
We call this net FCN-16s.
FCN-16s is learned end-to-end, initialized with the parameters of the last, coarser net, which we now call FCN-32s.
The new parameters acting on \texttt{pool4} are zero-initialized so that the net starts with unmodified predictions.
The learning rate is decreased by a factor of 100.

Learning this skip net improves performance on the validation set by 3.0 mean IU to 62.4.
Figure \ref{fig:evo} shows improvement in the fine structure of the output.
We compared this fusion with learning only from the \texttt{pool4} layer (which resulted in poor performance), and simply decreasing the learning rate without adding the extra link (which results in an insignificant performance improvement, without improving the quality of the output).

We continue in this fashion by fusing predictions from \texttt{pool3} with a $2\times$ upsampling of predictions fused from \texttt{pool4} and \texttt{conv7}, building the net FCN-8s.
We obtain a minor additional improvement to 62.7 mean IU, and find a slight improvement in the smoothness and detail of our output.
At this point our fusion improvements have met diminishing returns, both with respect to the IU metric which emphasizes large-scale correctness, and also in terms of the improvement visible e.g. in Figure \ref{fig:evo}, so we do not continue fusing even lower layers.

\begin{table}
\centering
\caption{
Comparison of skip \NAME s on a subset of PASCAL VOC2011 validation\protect\footnotemark[7].
Learning is end-to-end, except for FCN-32s-fixed, where only the last layer is fine-tuned. Note that FCN-32s is FCN-VGG16, renamed to highlight stride.
}
\begin{tabular}{r|cccc}
& \pixacc & \classacc & \meanIU & \fwIU \\
\hline
FCN-32s-fixed & 83.0 & 59.7 & 45.4 & 72.0 \\
FCN-32s & 89.1 & 73.3 & 59.4 & 81.4 \\
FCN-16s & 90.0 & 75.7 & 62.4 & 83.0 \\
FCN-8s & \textbf{90.3} & \textbf{75.9} & \textbf{62.7} & \textbf{83.2} \\
\end{tabular}
\label{tab:fcnstride}
\end{table}

\minisection{Refinement by other means}
Decreasing the stride of pooling layers is the most straightforward way to obtain finer predictions.
However, doing so is problematic for our VGG16-based net.
Setting the \texttt{pool5} layer to have stride 1 requires our convolutionalized \texttt{fc6} to have a kernel size of $14 \times 14$ in order to maintain its receptive field size.
In addition to their computational cost, we had difficulty learning such large filters.
We made an attempt to re-architect the layers above \texttt{pool5} with smaller filters, but were not successful in achieving comparable performance; one possible explanation is that the initialization from ImageNet-trained weights in the upper layers is important.

Another way to obtain finer predictions is to use the shift-and-stitch trick described in Section \ref{sec:shifting}.
In limited experiments, we found the cost to improvement ratio from this method to be worse than layer fusion.

\subsection{Experimental framework}
\label{sec:training}


\minisection{Optimization}
We train by SGD with momentum.
We use a minibatch size of 20 images and fixed learning rates of $10^{-3}$, $10^{-4}$, and $5^{-5}$ for FCN-AlexNet, FCN-VGG16, and FCN-GoogLeNet, respectively, chosen by line search.
We use momentum $0.9$, weight decay of $5^{-4}$ or $2^{-4}$, and doubled the learning rate for biases, although we found training to be insensitive to these parameters (but sensitive to the learning rate).
We zero-initialize the class scoring convolution layer, finding random initialization to yield neither better performance nor faster convergence.
Dropout was included where used in the original classifier nets.

\minisection{Fine-tuning}
We fine-tune all layers by back-propagation through the whole net.
Fine-tuning the output classifier alone yields only 70\% of the full fine-tuning performance as compared in Table \ref{tab:fcnstride}.
Training from scratch is not feasible considering the time required to learn the base classification nets.
(Note that the VGG net is trained in stages, while we initialize from the full 16-layer version.)
Fine-tuning takes three days on a single GPU for the coarse FCN-32s version, and about one day each to upgrade to the FCN-16s and FCN-8s versions.

\begin{figure}
\centering
\includegraphics[width=0.23\textwidth]{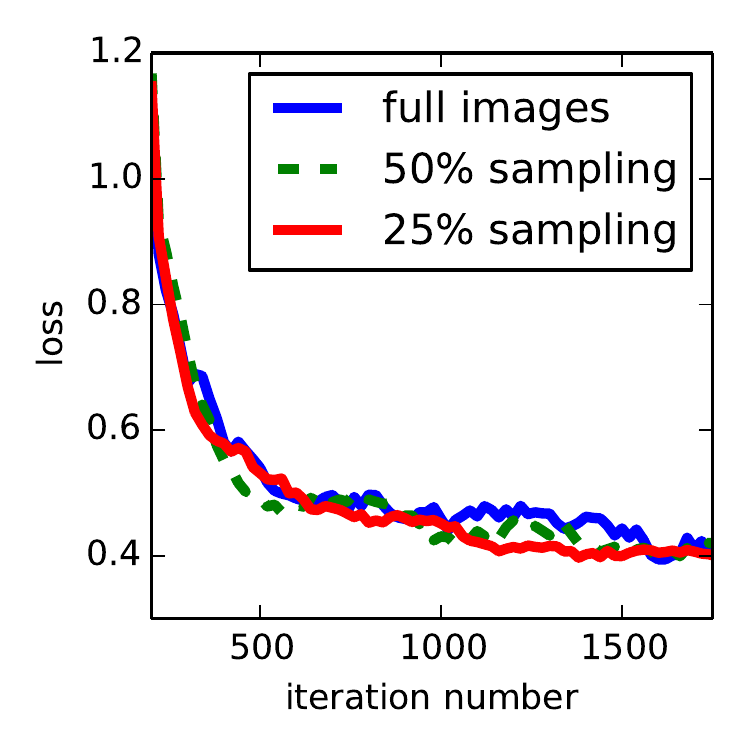}
\includegraphics[width=0.23\textwidth]{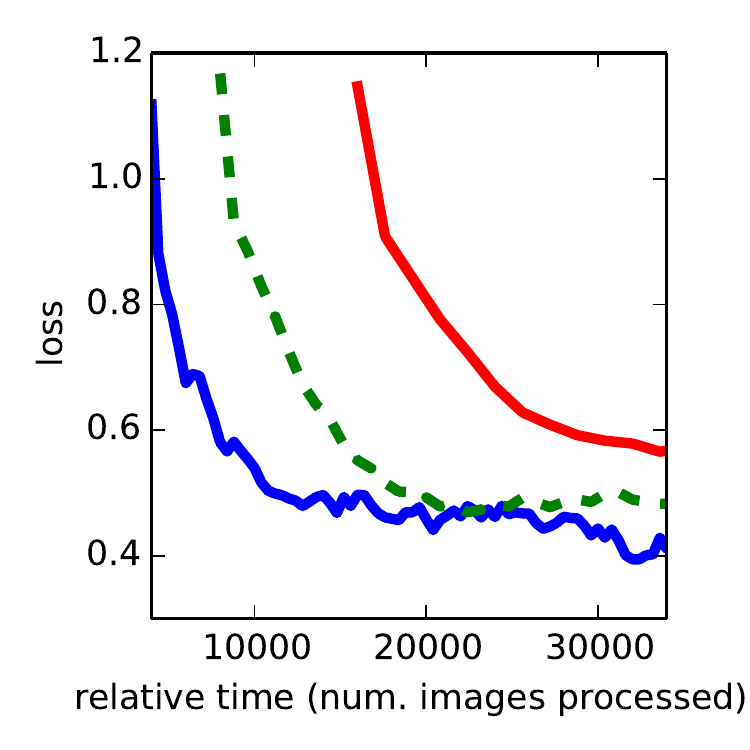}
\caption{
Training on whole images is just as effective as sampling patches, but results in faster (wall time) convergence by making more efficient use of data.
Left shows the effect of sampling on convergence rate for a fixed expected batch size, while right plots the same by relative wall time.
}
\label{fig:sampling}
\end{figure}

\minisection{Patch Sampling}
As explained in Section \ref{sec:patches}, our full image training effectively batches each image into a regular grid of large, overlapping patches.
By contrast, prior work randomly samples patches over a full dataset \cite{ning2005automatic, Ciresan, Farabet, Pinheiro, N4}, potentially resulting in higher variance batches that may accelerate convergence \cite{backprop}.
We study this tradeoff by spatially sampling the loss in the manner described earlier, making an independent choice to ignore each final layer cell with some probability $1 - p$.
To avoid changing the effective batch size, we simultaneously increase the number of images per batch by a factor $1/p$.
Note that due to the efficiency of convolution, this form of rejection sampling is still faster than patchwise training for large enough values of $p$ (e.g., at least for $p > 0.2$ according to the numbers in Section \ref{sec:adapting}).
Figure \ref{fig:sampling} shows the effect of this form of sampling on convergence.
We find that sampling does not have a significant effect on convergence rate compared to whole image training, but takes significantly more time due to the larger number of images that need to be considered per batch.
We therefore choose unsampled, whole image training in our other experiments.

\minisection{Class Balancing}
Fully convolutional training can balance classes by weighting or sampling the loss.
Although our labels are mildly unbalanced (about $3/4$ are background), we find class balancing unnecessary.

\minisection{Dense Prediction}
The scores are upsampled to the input dimensions by deconvolution layers within the net.
Final layer deconvolutional filters are fixed to bilinear interpolation, while intermediate upsampling layers are initialized to bilinear upsampling, and then learned.
Shift-and-stitch (Section \ref{sec:shifting}), or the filter rarefaction equivalent, are not used.

\minisection{Augmentation}
We tried augmenting the training data by randomly mirroring and ``jittering'' the images by translating them up to 32 pixels (the coarsest scale of prediction) in each direction.
This yielded no noticeable improvement.

\minisection{More Training Data}
The PASCAL VOC 2011 segmentation challenge training set, which we used for Table \ref{tab:base}, labels 1112 images.
Hariharan \etal \cite{BharathData} have collected labels for a much larger set of 8498 PASCAL training images, which was used to train the previous state-of-the-art system, SDS \cite{Bharath}.
This training data improves the FCN-VGG16 validation score%
\footnote{There are training images from \cite{BharathData} included in the PASCAL VOC 2011 val set, so we validate on the non-intersecting set of 736 images. An earlier version of this paper mistakenly evaluated on the entire val set.}
by 3.4 points to 59.4 mean IU.

\minisection{Implementation}
All models are trained and tested with Caffe \cite{caffe} on a single NVIDIA Tesla K40c.
The models and code will be released open-source on publication.

%


\section{Results}
\label{sec:results}

We test our FCN on semantic segmentation and scene parsing, exploring PASCAL VOC, NYUDv2, and SIFT Flow.
Although these tasks have historically distinguished between objects and regions, we treat both uniformly as pixel prediction.
We evaluate our FCN skip architecture%
\footnote{Our models and code are publicly available at \texttt{https://github.com/BVLC/caffe/wiki/Model-Zoo\#fcn}.}
on each of these datasets, and then extend it to multi-modal input for NYUDv2 and multi-task prediction for the semantic and geometric labels of SIFT Flow.

\minisection{Metrics}
We report four metrics from common semantic segmentation and scene parsing evaluations that are variations on pixel accuracy and region intersection over union (IU).
Let $n_{ij}$ be the number of pixels of class $i$ predicted to belong to class $j$, where there are $n_{\text{cl}}$ different classes, and let $t_i = \sum_j n_{ij}$ be the total number of pixels of class $i$. We compute:
\begin{itemize}[itemsep=0.2em,topsep=0pt,parsep=0pt,partopsep=0pt]
  \item pixel accuracy: \scalebox{0.8}{$\sum_i n_{ii} / \sum_i t_i$} 
  \item mean accuraccy: \scalebox{0.8}{$(1/n_{\text{cl}}) \sum_i n_{ii}/t_i$} 
  \item mean IU: \scalebox{0.8}{$(1/n_{\text{cl}}) \sum_i n_{ii} /\left(t_i + \sum_j n_{ji} - n_{ii}\right)$} 
  \item frequency weighted IU: \\\scalebox{0.8}{$\left(\sum_k t_k\right)^{-1} \sum_i t_i n_{ii} /\left(t_i + \sum_j n_{ji} - n_{ii}\right)$} 
\end{itemize}

\minisection{PASCAL VOC}
Table \ref{tab:pascaltest} gives the performance of our FCN-8s on the test sets of PASCAL VOC 2011 and 2012, and compares it to the previous state-of-the-art, SDS \cite{Bharath}, and the well-known R-CNN \cite{RCNN}.
We achieve the best results on mean IU\footnote{This is the only metric provided by the test server.} by a relative margin of 20\%.
Inference time is reduced $114\times$ (convnet only, ignoring proposals and refinement) or $286\times$ (overall).

\begin{table}[h!]
\caption{
Our fully convolutional net gives a 20\% relative improvement over the state-of-the-art on the PASCAL VOC 2011 and 2012 test sets, and reduces inference time.
}
\scalebox{0.94}{
\begin{tabular}{r|ccc}
& mean IU & mean IU & inference \\
& VOC2011 test & VOC2012 test & time \\
\hline
R-CNN \cite{RCNN} & 47.9 & - & - \\
SDS \cite{Bharath} & 52.6 & 51.6 & $\sim$ 50 s \\
FCN-8s & \textbf{62.7} & \textbf{62.2} & \bf{$\sim$ 175 ms}
\end{tabular}
}
\label{tab:pascaltest}
\end{table}

\textbf{NYUDv2} \cite{NYUDv2} is an RGB-D dataset collected using the Microsoft Kinect.
It has 1449 RGB-D images, with pixelwise labels that have been coalesced into a 40 class semantic segmentation task by Gupta \etal \cite{Saurabh1}.
We report results on the standard split of 795 training images and 654 testing images.
(Note: all model selection is performed on PASCAL 2011 val.)
Table \ref{tab:nyud} gives the performance of our model in several variations.
First we train our unmodified coarse model (FCN-32s) on RGB images.
To add depth information, we train on a model upgraded to take four-channel RGB-D input (early fusion).
This provides little benefit, perhaps due to the difficultly of propagating meaningful gradients all the way through the model.
Following the success of Gupta \etal \cite{Saurabh}, we try the three-dimensional HHA encoding of depth, training nets on just this information, as well as a ``late fusion'' of RGB and HHA where the predictions from both nets are summed at the final layer, and the resulting two-stream net is learned end-to-end.
Finally we upgrade this late fusion net to a 16-stride version.

\textbf{SIFT Flow} is a dataset of 2,688 images with pixel labels for 33 semantic categories (``bridge'', ``mountain'', ``sun''), as well as three geometric categories (``horizontal'', ``vertical'', and ``sky'').
An FCN can naturally learn a joint representation that simultaneously predicts both types of labels.
We learn a two-headed version of FCN-16s with semantic and geometric prediction layers and losses.
The learned model performs as well on both tasks as two independently trained models, while learning and inference are essentially as fast as each independent model by itself.
The results in Table \ref{tab:siftflow}, computed on the standard split into 2,488 training and 200 test images,%
\footnote{Three of the SIFT Flow categories are not present in the test set.
We made predictions across all 33 categories, but only included categories actually present in the test set in our evaluation.
(An earlier version of this paper reported a lower mean IU, which included all categories either present or predicted in the evaluation.)}
show state-of-the-art performance on both tasks.

\begin{table}
\centering
\begin{minipage}{0.45\textwidth}
\caption{
Results on NYUDv2.
\textit{RGBD} is early-fusion of the RGB and depth channels at the input.
\textit{HHA} is the depth embedding of \cite{Saurabh} as horizontal disparity, height above ground, and the angle of the local surface normal with the inferred gravity direction.
\textit{RGB-HHA} is the jointly trained late fusion model that sums RGB and HHA predictions.
}
\begin{tabular}{r|cccc}
  & \pixacc & \classacc & \meanIU & \fwIU \\
\hline
Gupta \etal \cite{Saurabh} & 60.3 & - & 28.6 & 47.0 \\
FCN-32s RGB & 60.0 & 42.2 & 29.2 & 43.9 \\
FCN-32s RGBD & 61.5 & 42.4 & 30.5 & 45.5 \\
FCN-32s HHA & 57.1 & 35.2 & 24.2 & 40.4 \\
FCN-32s RGB-HHA & 64.3 & 44.9 & 32.8 & 48.0 \\
FCN-16s RGB-HHA & \textbf{65.4} & \textbf{46.1} & \textbf{34.0} & \textbf{49.5}
\end{tabular}
\label{tab:nyud}
\end{minipage}
\end{table}
\begin{table}
\begin{minipage}{0.45\textwidth}
\caption{
Results on SIFT Flow\protect\footnotemark[10] with class segmentation (center) and geometric segmentation (right).
Tighe \cite{Superparsing} is a non-parametric transfer method.
Tighe 1 is an exemplar SVM while 2 is SVM + MRF.
Farabet is a multi-scale convnet trained on class-balanced samples (1) or natural frequency samples (2).
Pinheiro is a multi-scale, recurrent convnet, denoted {\sc r}CNN$_3$ $(\circ^3)$.
The metric for geometry is pixel accuracy.
}
\scalebox{0.95}{
\begin{tabular}{r|cccc|c}
& \pixacc & \classacc & \meanIU & \fwIU & \parbox[b]{0.3in}{geom.\\acc.} \\
\hline
Liu \etal \cite{sift-flow} & 76.7 & - & - & - & - \\
Tighe \etal \cite{Superparsing} & - & - & - & - & 90.8 \\
Tighe \etal \cite{tighe2013finding} 1 & 75.6 & 41.1 & - & - & - \\
Tighe \etal  \cite{tighe2013finding} 2 & 78.6 & 39.2 & - & - & - \\
Farabet \etal \cite{Farabet} 1 & 72.3 & 50.8 & - & - & - \\
Farabet \etal \cite{Farabet} 2 & 78.5 & 29.6 & - & - & - \\
Pinheiro \etal \cite{Pinheiro} & 77.7 & 29.8 & - & - & - \\
FCN-16s & \textbf{85.2} & \textbf{51.7} & 39.5 & 76.1 & \textbf{94.3}
\end{tabular}
}
\label{tab:siftflow}
\end{minipage}
\end{table}

\begin{figure}
\centering
\setlength{\tabcolsep}{1pt}
\scalebox{0.95}{
\begin{tabular}{cccc}
FCN-8s & SDS \cite{Bharath} & Ground Truth & Image \\
\includegraphics[width=0.12\textwidth]{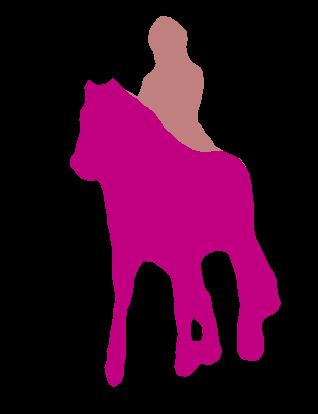} &
\includegraphics[width=0.12\textwidth]{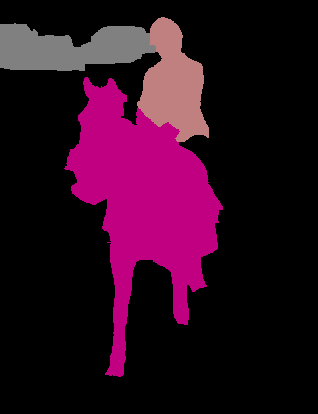} &
\includegraphics[width=0.12\textwidth]{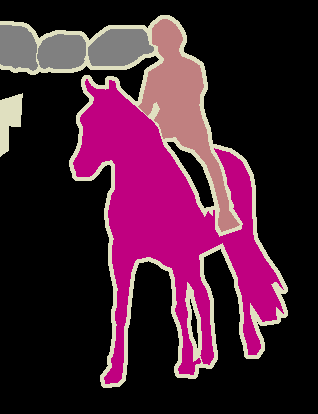} &
\includegraphics[width=0.12\textwidth]{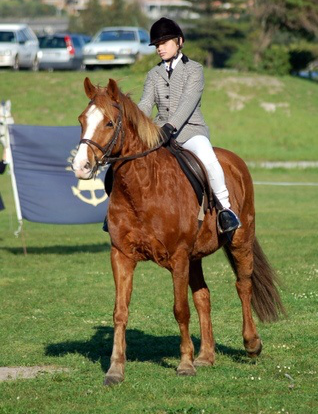} \\
\includegraphics[width=0.12\textwidth]{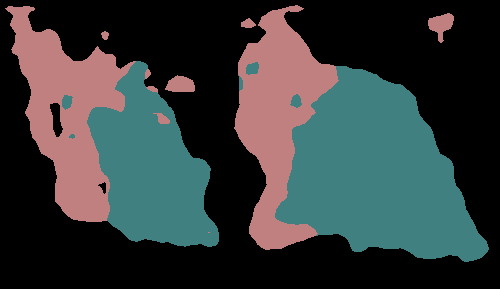} &
\includegraphics[width=0.12\textwidth]{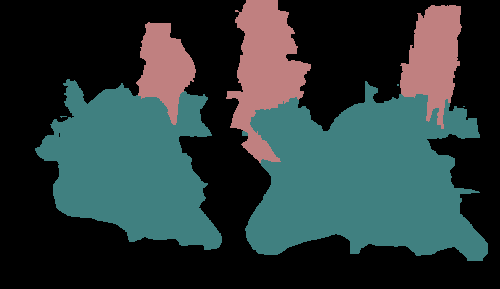} &
\includegraphics[width=0.12\textwidth]{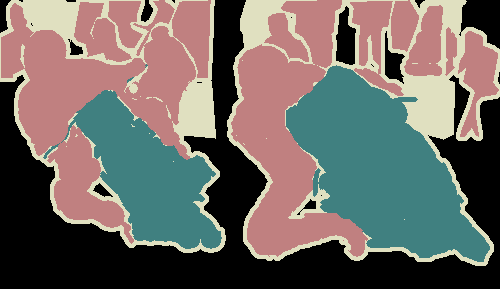} &
\includegraphics[width=0.12\textwidth]{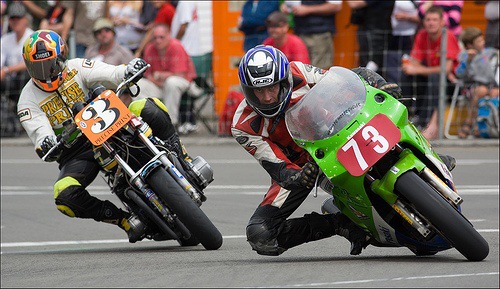} \\
\includegraphics[width=0.12\textwidth]{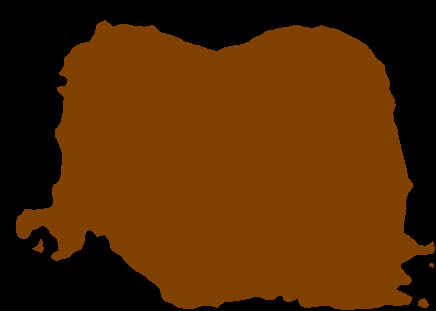} &
\includegraphics[width=0.12\textwidth]{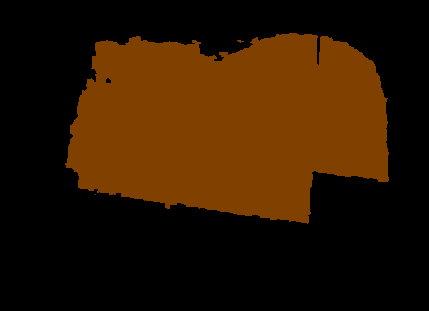} &
\includegraphics[width=0.12\textwidth]{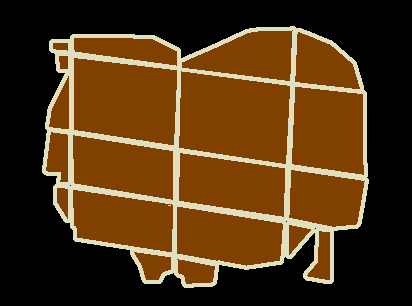} &
\includegraphics[width=0.12\textwidth]{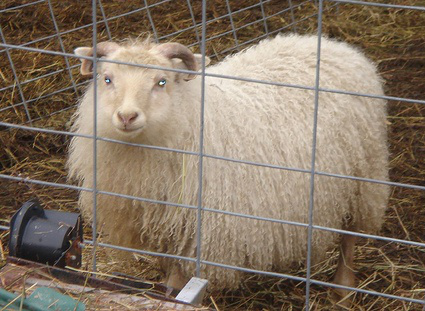} \\
\includegraphics[width=0.12\textwidth]{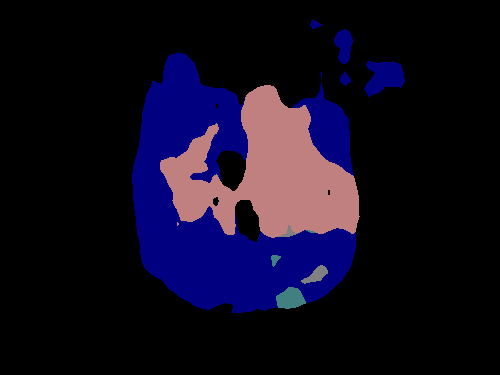} &
\includegraphics[width=0.12\textwidth]{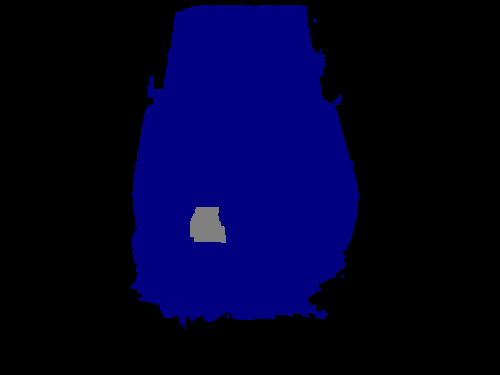} &
\includegraphics[width=0.12\textwidth]{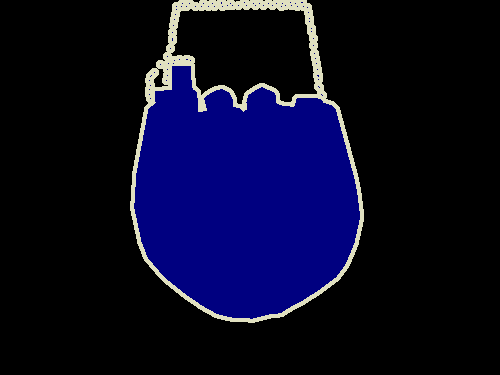} &
\includegraphics[width=0.12\textwidth]{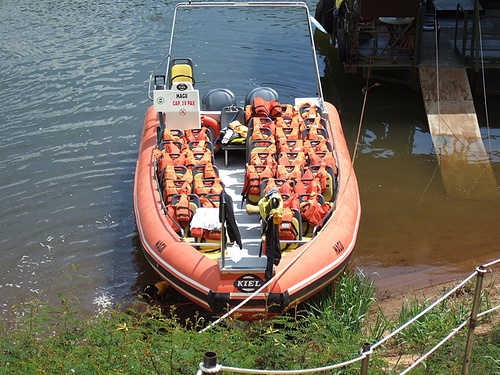} \\
\end{tabular}
}
\caption{
Fully convolutional segmentation nets produce state-of-the-art performance on PASCAL.
The left column shows the output of our highest performing net, FCN-8s.
The second shows the segmentations produced by the previous state-of-the-art system by Hariharan \etal \cite{Bharath}.
Notice the fine structures recovered (first row), ability to separate closely interacting objects (second row), and robustness to occluders (third row).
The fourth row shows a failure case: the net sees lifejackets in a boat as people.
}
\end{figure}



\section{Conclusion}

\Name s are a rich class of models, of which modern classification convnets are a special case.
Recognizing this, extending these classification nets to segmentation, and improving the architecture with multi-resolution layer combinations dramatically improves the state-of-the-art, while simultaneously simplifying and speeding up learning and inference.

\ifcvprfinal
\minisection{Acknowledgements}
This work was supported in part by DARPA's MSEE and SMISC programs, NSF awards IIS-1427425, IIS-1212798, IIS-1116411, and the NSF GRFP, Toyota, and the Berkeley Vision and Learning Center.
We gratefully acknowledge NVIDIA for GPU donation.
We thank Bharath Hariharan and Saurabh Gupta for their advice and dataset tools.
We thank Sergio Guadarrama for reproducing GoogLeNet in Caffe.
We thank Jitendra Malik for his helpful comments.
Thanks to Wei Liu for pointing out an issue wth our SIFT Flow mean IU computation and an error in our frequency weighted mean IU formula.
\fi

\appendix
\section{Upper Bounds on IU}
\label{sec:ub}

In this paper, we have achieved good performance on the mean IU segmentation metric even with coarse semantic prediction.
To better understand this metric and the limits of this approach with respect to it, we compute approximate upper bounds on performance with prediction at various scales.
We do this by downsampling ground truth images and then upsampling them again to simulate the best results obtainable with a particular downsampling factor.
The following table gives the mean IU on a subset of PASCAL 2011 val for various downsampling factors.

\begin{center}
\begin{tabular}{r|l}
factor & mean IU \\
\hline
128 & 50.9 \\
64 & 73.3 \\
32 & 86.1 \\
16 & 92.8 \\
8  & 96.4 \\
4  & 98.5
\end{tabular}
\end{center}

Pixel-perfect prediction is clearly not necessary to achieve mean IU well above state-of-the-art, and, conversely, mean IU is a not a good measure of fine-scale accuracy.

\section{More Results}
\label{sec:context}

We further evaluate our FCN for semantic segmentation.

\textbf{PASCAL-Context} \cite{mottaghi2014role} provides whole scene annotations of PASCAL VOC 2010.
While there are over 400 distinct classes, we follow the 59 class task defined by \cite{mottaghi2014role} that picks the most frequent classes.
We train and evaluate on the training and val sets respectively.
In Table \ref{tab:pascal-context}, we compare to the joint object + stuff variation of Convolutional Feature Masking \cite{CFM} which is the previous state-of-the-art on this task.
FCN-8s scores 35.1 mean IU for an 11\% relative improvement.

\begin{table}[h]
\centering
\caption{
Results on PASCAL-Context.
\textit{CFM} is the best result of \cite{CFM} by convolutional feature masking and segment pursuit with the VGG net.
\textit{O$_2$P} is the second order pooling method \cite{O2P} as reported in the \emph{errata} of \cite{mottaghi2014role}.
The 59 class task includes the 59 most frequent classes while the 33 class task consists of an easier subset identified by \cite{mottaghi2014role}.
}
\begin{tabular}{r|cccc}
59 class  & \pixacc & \classacc & \meanIU & \fwIU \\
\hline
O$_2$P  & - & - & 18.1 & - \\
CFM     & - & - & 31.5 & - \\
FCN-32s & 63.8 & 42.7 & 31.8 & 48.3 \\
FCN-16s & 65.7 & 46.2 & 34.8 & 50.7 \\
FCN-8s  & \textbf{65.9} & \textbf{46.5} & \textbf{35.1} & \textbf{51.0} \\
& & & & \\
33 class  &  &  &  & \\
\hline
O$_2$P  & - & - & 29.2 & - \\
CFM     & - & - & 46.1 & - \\
FCN-32s & 69.8 & 65.1 & 50.4 & 54.9 \\
FCN-16s & \textbf{71.8} & \textbf{68.0} & 53.4 & 57.5 \\
FCN-8s & \textbf{71.8} & 67.6 & \textbf{53.5} & \textbf{57.7} \\
\end{tabular}
\label{tab:pascal-context}
\end{table}

\section*{Changelog}

The arXiv version of this paper is kept up-to-date with corrections and additional relevant material.
The following gives a brief history of changes.
\\

\noindent \minisection{v2}
Add Appendix \ref{sec:ub} giving upper bounds on mean IU and Appendix \ref{sec:context} with PASCAL-Context results.
Correct PASCAL validation numbers (previously, some val images were included in train), SIFT Flow mean IU (which used an inappropriately strict metric), and an error in the frequency weighted mean IU formula.
Add link to models and update timing numbers to reflect improved implementation (which is publicly available).

{\small
\bibliographystyle{ieee}
\bibliography{bib}
}

\end{document}